\documentclass{article}

\usepackage{arxiv}

\usepackage[utf8]{inputenc} 
\usepackage[T1]{fontenc}    
\usepackage{hyperref}       
\usepackage{url}            
\usepackage{booktabs}       
\usepackage{amsfonts}       
\usepackage{nicefrac}       
\usepackage{microtype}      
\usepackage{lipsum}		
\usepackage{graphicx}
\usepackage{natbib}
\usepackage{doi}
\usepackage{placeins}
\usepackage{palatino,epsfig,latexsym}
\usepackage[acronym]{glossaries}
\usepackage{bm}

\DeclareMathOperator*{\argmin}{arg\,min}

\makeglossaries
\newacronym{FDA}{FDA}{full domain analysis}
\newacronym{QD}{QD}{quality diversity}
\newacronym{MMO}{MMO}{multimodal optimization}
\newacronym{MOO}{MOO}{multiobjective optimization}
\newacronym{BO}{BO}{Bayesian optimization}
\newacronym{GP}{GP}{Gaussian process model}
\newacronym{FFD}{FFD}{free form deformation}
\newacronym{CPPN}{CPPN}{compositional pattern producing networks}
\newacronym{SAO}{SAO}{surrogate-assisted optimization}
\newacronym{NSGA-II}{NSGA-II}{non-dominated sorting genetic algorithm}
\newacronym{GM}{GM}{generative model}
\newacronym{VAE}{VAE}{variational autoencoder}
\newacronym{GAN}{GAN}{generative adversarial network}
\newacronym{GA}{GA}{genetic algorithm}
\newacronym{NURBS}{NURBS}{non-uniform rational b-Splines}
\newacronym{CFD}{CFD}{computational fluid dynamics}
\newacronym{SPHEN}{SPHEN}{surrogate-assisted phenotypic niching}
\newacronym{LBM}{LBM}{lattice Boltzmann method}
\newacronym{EGO}{EGO}{efficient global optimisation}
\title{Full domain analysis in fluid dynamics}

\date{May 19, 2025}	

\author{ 
\href{https://orcid.org/0000-0002-8668-1796}{\includegraphics[scale=0.06]{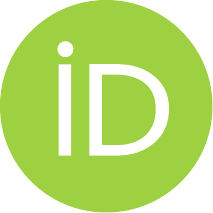}\hspace{1mm}Alexander~Hagg}\\
Institute of Technology, Resource and Energy-efficient Engineering (TREE) \\
Bonn-Rhein-Sieg University of Applied Sciences \\
Sankt Augustin, 53757, Germany \\
\texttt{alexander.hagg@h-brs.de} \\
\And
\href{https://orcid.org/0000-0002-4632-0929}{\includegraphics[scale=0.06]{orcid.pdf}\hspace{1mm}Adam~Gaier}\\
Autodesk Research, London, UK \\
\And
\href{https://orcid.org/0000-0003-3263-7287}{\includegraphics[scale=0.06]{orcid.pdf}\hspace{1mm}Dominik~Wilde}\\
Institute of Technology, Resource and Energy-efficient Engineering (TREE) \\
Bonn-Rhein-Sieg University of Applied Sciences \\
Sankt Augustin, 53757, Germany \\
\And
\href{https://orcid.org/0000-0003-1133-9424}{\includegraphics[scale=0.06]{orcid.pdf}\hspace{1mm}Alexander~Asteroth}\\
Institute of Technology, Resource and Energy-efficient Engineering (TREE) \\
Bonn-Rhein-Sieg University of Applied Sciences \\
Sankt Augustin, 53757, Germany \\
\texttt{alexander.asteroth@h-brs.de} \\
\And
\href{https://orcid.org/0000-0003-2056-6960}{\includegraphics[scale=0.06]{orcid.pdf}\hspace{1mm}Holger~Foysi}\\
Chair of Fluid Mechanics, University of Siegen, Germany \\
\texttt{holger.foysi@uni-siegen.de} \\
\And
\href{https://orcid.org/0000-0003-1480-6745}{\includegraphics[scale=0.06]{orcid.pdf}\hspace{1mm}Dirk~Reith}\\
Institute of Technology, Resource and Energy-efficient Engineering (TREE) \\
Bonn-Rhein-Sieg University of Applied Sciences \\
Sankt Augustin, 53757, Germany \\
Fraunhofer Institute for Algorithms and Scientific Computing (SCAI), Sankt Augustin, Germany\\
\texttt{dirk.reith@h-brs.de} \\
}

\renewcommand{\shorttitle}{Full domain analysis in fluid dynamics}

\hypersetup{
pdftitle={Full domain analysis in fluid dynamics},
pdfsubject={cs.ML},
pdfauthor={Ludovico Scarton, Alexander Hagg},
pdfkeywords={multi-solution optimization, machine learning, efficient domain analysis, problem diversity, generative methods, fluid dynamics},
}

\begin{document}
\maketitle

\begin{abstract}
	Novel techniques in evolutionary optimization, simulation and machine learning allow for a broad analysis of domains like fluid dynamics, in which computation is expensive and flow behavior is complex.
	Under the term of full domain analysis we understand the ability to efficiently determine the full space of solutions in a problem domain, and analyze the behavior of those solutions in an accessible and interactive manner. The goal of full domain analysis is to deepen our understanding of domains by generating many examples of flow, their diversification, optimization and analysis.
	We define a formal model for full domain analysis, its current state of the art, and requirements of sub-components.
	Finally, an example is given to show what we can learn by using full domain analysis.
	Full domain analysis, rooted in optimization and machine learning, can be a helpful tool in understanding complex systems in computational physics and beyond.
\end{abstract}




\maketitle

\section{Introduction}
\label{sec:introduction}
Problem solving in fluid dynamics is a major research and development field with a large impact on our energy usage as well as a large potential to make our society more sustainable.
Due to the complexity of the field, creativity and innovation can be incredibly hard.
Only by increasing our understanding of entire problem domains can we systematically discover innovative solutions and reduce the risk involved in early decision making. 
Our algorithms need to express this need for innovation and risk reduction. 

This work therefore introduces \textit{\glsfirst{FDA}}: the ability to efficiently understand the full space of solutions (e. g. shape designs) in a problem domain, and analyze the behavior of those solutions in an accessible and interactive manner.
\gls{FDA} does not refer to a specific method but rather to a methodological framework that specifies which components are required when analyzing expensive domains in a structured fashion. 
Applying \gls{FDA} to problem domains in fluid dynamics requires taking into account the particularities of flow behavior and inefficiencies of fluid simulations. 
Small changes in flow environments can have significant and non-local effects on the airflow. 
If we want to understand these effects, we need efficient methods and frameworks to help us to predict relevant flow features in a large diversity of settings (e. g. shapes).
We also need to be able to represent this understanding in a concise manner, to enable us to handle the mental load on engineers analyzing such complex domains.

Assuming engineers are only interested in solutions that are optimal w. r. t. some measure, the following questions must be answered to make this a reality:

\begin{enumerate}
	\item How should designs be encoded to allow for efficiency in both determining solution quality and diversity of solutions?
	\item How can a diverse set of solutions be created algorithmically?
	\item How can a large set of solutions be created efficiently?
	\item How can large and diverse sets of solutions and flow artifacts be visualized effectively so users can understand and navigate the design space?
\end{enumerate}

In the field of artificial intelligence, specifically optimization and machine learning, a new body of work has emerged that allows automating creative processes that were classically only possible within the capabilities of the human mind. 
Methods like deep generative models \citep[see][]{wang2021airfoil} and divergent optimization \citep[see][]{gaier2018data} now exist that enable generating large sets of innovative solutions.
The ability to generate many different, high-quality solutions to a problem domain in an efficient manner has given us a big step towards \gls{FDA} and understanding and reducing unwanted consequences of design decisions.

In this work we explore synergistic effects of state-of-the-art evolutionary optimization, surrogate modeling techniques, and deep learning, applied to the automated design of structures in the domain of fluid dynamics. 
We present an \gls{FDA} framework and tool set that allows us to better explore and understand fluid dynamics design and develop automated design methods.

\begin{figure}
	\centering
	\includegraphics[width=0.7\textwidth]{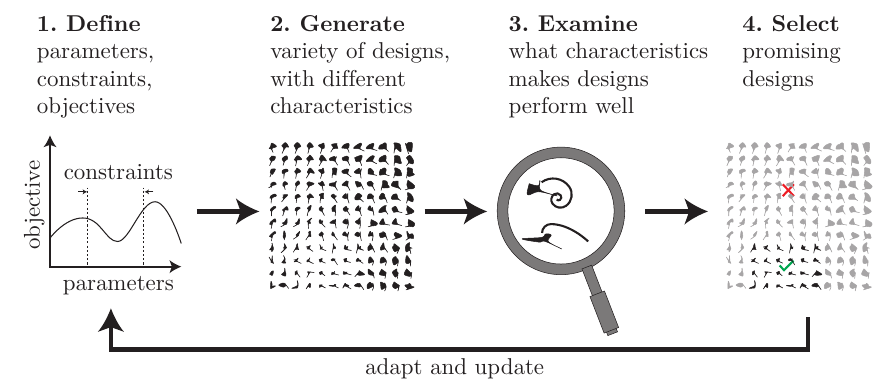}
	\caption{User process perspective on \gls{FDA}. After the user defines the domain through available initial parameters, constraints and objectives, the prior (1), a large variety of designs is automatically generated (2). The user examines what makes designs perform well (3). They then select promising design regions which allows \gls{FDA} to update the initial domain definition (4). The next iteration of the \gls{FDA} process zooms in on the user's region of interest.}
	\label{fig:framework}
\end{figure}

Fig.~\ref{fig:framework} shows the \gls{FDA} framework through the perspective of the user process.
The goal of \gls{FDA} is to get a representative set of flow features and flow manifestations around a large amount of possible shapes in a predefined environment.
It is defined as an iterative process that is started by the user defining parameters, constraints, and objectives that define the problem domain (1). 
Based on this, an efficient generative algorithm then creates a large variety of designs (2). 
The user then examines the design performance and characteristics (3) and can decide to select promising design regions, or design classes, (4). 
Based on this selection, parameters, constraints and objectives can be adapted, after which the \gls{FDA} process starts its next iteration. 
This allows a user to quickly get an overview over what designs perform well and zoom in on or recombine design regions.

Machine-learning techniques are used to guide iterative experimentation with novel designs. 
The major contribution of this work is to provide a framework that describes the goal and requirement set of \gls{FDA} and its components (encodings\footnote{The manner in which we encode, or parameterize the configuration of a flow domain.}, search, \gls{CFD}, efficiency and visualization). 
We provide examples of state-of-the-art methods that are able to fulfill these requirements. 
The following \gls{FDA} components are discussed in this work:
\begin{enumerate}
	\item encoding of solutions,
	\item effective and efficient divergent search,
	\item fast \gls{CFD} solver that supports accurate flow for diverse shapes,
	\item statistical learning methods to efficiently sample and predict characteristics of solutions,
	\item machine learning methods that learn characteristics and representations of solution sets.
\end{enumerate}

The examples of \gls{FDA} components can be used independently but all belong to an ecosystem of \gls{FDA} methods. 
Most of this article gives examples of components and their application to the domain of shape optimization in fluid dynamics.
We also elaborate on their shortfalls and future work arising from the requirements of \gls{FDA}.
The next sections discuss novel methods that, together, solve the problems mentioned here. 
Section~\ref{sec:encodings} specifies free form as well as data-driven shape encodings that provide the search space. 
The encodings are used to find diverse solution sets, described in Section~\ref{sec:search}. 
Section~\ref{sec:simulation} describes the requirements of simulation environments when evaluating diverse shape sets.
The matter of algorithmic efficiency, a key element of \gls{FDA}, is considered  in Section~\ref{sec:efficiency}.
A fully implemented example of what can be achieved with \gls{FDA} is given in Section \ref{sec:example}
Finally, we provide recommendations about future work involving \gls{FDA} methods in Section~\ref{sec:conclusions}.




\section{Encodings}
\label{sec:encodings}
A generative design system requires an in silico representation of the design problem and its prospective solutions. 
In the context of the work herein, this requires encoding shapes. 
Several intuitive encodings exist, all with their own benefits and problems. 
It is tempting directly encode the coordinates of such a free shape directly into the search space, as this can result in any shapes. 
However, this approach would require a large number of parameters, increasing the dimensionality of the search space, and produce many invalid shapes due to intersecting outlines. 
One can circumvent these problems by parameterizing certain characteristics of a basic shape, which was derived, for instance, in design optimization of airfoils \citep[see][]{jacobs1933characteristics}. 
However, such basic shapes might not exist for many areas of design where components are customized to create mechanisms that meet specific path and force characteristics. 
Therefore, there are several things to consider when choosing a shape representation. 
This research emphasizes search space dimension, coverage, validity, and navigation of encodings.
In what follows we define requirements of encodings and discuss whether and how common encodings fulfill these requirements. 

\subsection{Requirements}
\label{sec:encodings:requirements}
\begin{figure}
	\centering
	\includegraphics[width=1\linewidth]{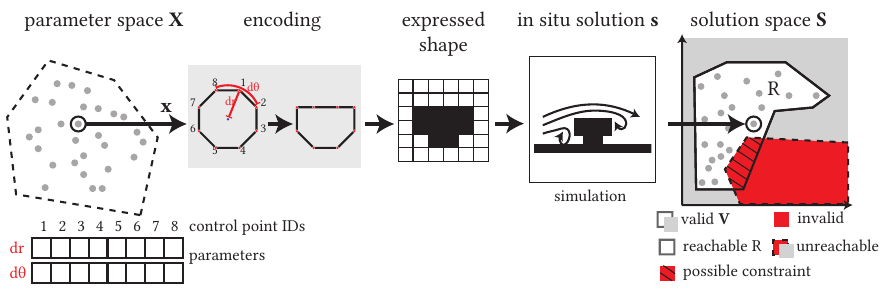}
	\caption{While search and optimization takes place in which we call the parameter space $\mathbf{X}$, through the encoding's expression and its in-situ simulation, only a part of the solution space $\mathbf{S}$ can be reached, the reachable manifold $\mathbf{R}$. This manifold might include invalid solutions (necessitating constraints on the parameter space). The goal is to maximize $\mathbf{R}$ w. r. t. the valid subspace $\mathbf{V}$.}
	\label{fig:surface}
\end{figure}

In optimization, solutions are usually encoded by a number of real valued parameters, i. e. parameter space $\mathbf{X}$ (see figure~\ref{fig:surface}).
$\mathbf{x} \in \mathbf{X}$ is a vector that is transformed into the final shape $\mathbf{s}$ by some expression method $e$, which we will discuss hereafter.
The full solution space $\mathbf{S}$ represents the vector space containing all possible solutions.
This separation between $\mathbf{X}$, therein called a genetic space, and a phenotypic space $\mathbf{S}$, is common and explicitly used in evolutionary computation \citep[see][]{holland1992adaptation}.
By independently treating these spaces, we can more easily investigate the various encodings, the translation between $\mathbf{X}$ and $\mathbf{S}$, and the encodings' constraints.
For fluid flow around obstacles or shapes, these shapes are elements taken from the solution space $\mathbf{S}$, subject to high spatial (and possibly temporal) resolution. 
Therefore, $\mathbf{S}$ is usually of much higher dimensionality than the parameter space $\mathbf{X}$.
Encodings aim to reduce this dimensionality with a function that maps a low-dimensional search space $\mathbf{X}$ to a reachable manifold $\mathbf{R}$ in $\mathbf{S}$ with $dim(\mathbf{R}) \ll dim(\mathbf{S})$.
Only points on $\mathbf{R}$ can be reached by points in $\mathbf{X}$.

In the following we define requirements to and examples of encodings, describe their strengths and weaknesses and make recommendations for further research. 

\subsubsection{Reachability}
\textit{Create a wide variety of designs, maximizing the reachable region, or its coverage of valid solution space $\mathbf{V}$.} 
The quality of an \gls{FDA} method relies on the ability to reach as many solutions as possible.
The reachable solution space $\mathbf{R}$ contains all reachable in situ solutions.
$\mathbf{S}$ contains all possible valid solutions and invalid solutions.
The goal is to maximize $\mathbf{R}$ w. r. t. the solution space $\mathbf{S}$.
If the encoding itself cannot produce certain solutions that are valid and desirable, due to misalignment between what a user wants and how they formalized the encoding, \gls{FDA} will systematically miss those desirable solutions. 
The requirement on \textit{reachability} or completeness \citep[see][]{olhofer2000optimisation} aims for the encoding to guarantee maximal degree of freedom in order to avoid unnecessary constraints on the phenotype space.

\subsubsection{Validity}

\textit{Minimize the number of invalid solutions on $\mathbf{R}$.}
Often, a number of reachable and/or unreachable solutions are also invalid, e. g. when these shapes can or should not be manufactured due to limitations in machining or design.
Although we want to understand invalid solutions as well, we mostly want to be able to avoid reaching them. 
The number of invalid solutions in $\mathbf{S}$ can be much greater than $\mathbf{V}$.
It is not possible to evaluate invalid solutions and so we cannot calculate their fitness value.
Hence, the navigation through the search space $\mathbf{X}$ would be impossible or at least ineffective \citep[see][]{lapok2020evolving}.
In \gls{FDA}, a primary goal of an encoding is to maximize the reachable manifold $\mathbf{R}$ w. r. t. the valid subspace $\mathbf{V}$, which can contain multiple unconnected regions.
To ensure that the reachable manifold $\mathbf{R}$ does not contain any invalid solutions, constraints can be applied to $\mathbf{X}$.
Adding constraints can be tedious and usually needs multiple iterations, depending on the complexity of the encoding's mapping.
It is preferred that the encoding finds a manifold that contains as many valid and as few invalid solutions as possible, without the need for constraints.

\subsubsection{Searchability}
\textit{Minimize dimensionality, ambiguous w. r. t. sensitivity}
The encoding provides the search space $\mathbf{X}$ for optimization.
The solution space $\mathbf{S}$ suffers from the curse of dimensionality, a term coined by \citet{bellman1966dynamic}.
It dictates that the relative difference of the distances of the closest and farthest data points goes to zero as the dimensionality increases. 
This phenomenon already occurs at low numbers of dimensions, which was shown by \citet{beyer1999nearest}. 
We therefore prefer compactness \citep[see][]{olhofer2000optimisation} of $\mathbf{X}$ to minimize its dimensionality and increase search performance.

According to authors like \citet{olhofer2000optimisation}, small steps in $\mathbf{X}$ should lead to small steps in $\mathbf{S}$. 
Encodings that disobey this rule are coined sensitive \citep[see][]{hagg2021discovering}.
However, sensitivity  of an encoding can help finding diverse shapes by allowing a search algorithm to jump around in $\mathbf{S}$ and find disconnected high-fitness regions. 
Making use of lower-performing solutions as stepping stones towards better solutions, some evolutionary optimization algorithms are able to produce high diversity solution sets \citep[see][]{nordmoen2021map}.
So while the dimensionality of $\mathbf{X}$ should be kept low, it is unclear, whether sensitivity is preferred or not in divergent search.
Some evidence exists that it might be beneficial in the context of \gls{FDA}.

\subsubsection{Predictability}
\textit{Allow for efficient search using predictive models.}
Efficient search in the fluid dynamics domain usually encompasses predictive models that allows us to replace some expensive evaluations, i. e. the use of a \gls{CFD} solver. 
The encoding itself needs to be suitable for use with predictive (machine learning) models. This is done by learning to predict a function $\mathbf{x} \rightarrow f(\mathbf{x})$, where $f$ is a fitness function that determines the quality of a solution based on its parameters $x$. 
Sensitive encodings (i. e. small steps in $\mathbf{X}$ produce large steps in $\mathbf{S}$) are usually harder to predict accurately, as large steps in $\mathbf{S}$ can entail large qualitative changes. 
We do point out that in this case, large steps might still be easy to model, for monotonous transformations, e.g. when a large step just means that the shape is increased in size. 
Especially when the air flow changes in a qualitative manner, the modeling problem becomes more difficult, requiring more samples from simulation, more complex models or slower model conversion.
Local models, e.g. those that are commonly used in optimization, might become too complex if used in \gls{FDA}.
Hence, it might be more beneficial to use models that approximate the entire design space.

\subsubsection{Human understanding and effort}
\textit{Allow engineers to understand the domain and the implications of their definition. Minimize effort to define domain. Minimize effort to use results in production.}
Domains in fluid dynamics take a lot of effort to formally define and code.
In classical optimization, e. g. airfoil or wing design, the variety of possible shapes is often confined around a prior of design decisions.
These design decisions are part of the encoding formalization process.
In \gls{FDA} for fluid dynamics we aim to enhance the engineer's intuition, i. e. maximize the diversity of solutions and enhance the engineer's understanding of how morphological and flow features are correlated.
We therefore might prefer encodings that have more degrees of freedom and a larger range of possible solutions $\mathbf{s} \in \mathbf{S}$ than what is common in fluid dynamics optimization. 
Finding solutions on paper is but the first step towards realizing them in the real world.
Solutions that are defined in industry-compatibly formats could be preferred.
The encodings should allow engineers to understand, fine tune and realize in the real world.

\subsubsection{Prior examples}
\textit{Support learning from prior examples.}
In large and complex domains, we might want to be able to insert prior knowledge about known working solutions to kick start \gls{FDA}.

\subsection{State of the Art}
\begin{figure}
	\centering
	\includegraphics[width=1\linewidth]{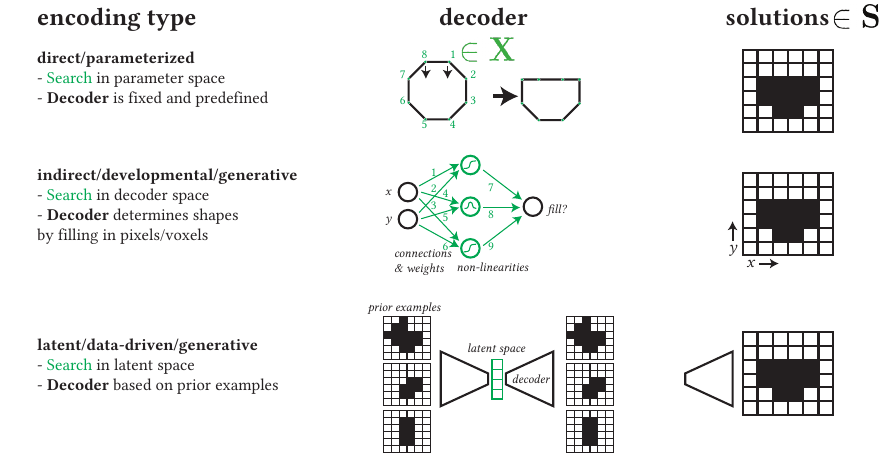}
	\caption{Three main categories of shape encodings to produce solutions in $\mathbf{S}$. \textit{Direct, parameterized encodings} use a manually defined decoder that determines spline shapes. \textit{Indirect encodings} search the shapes indirectly, by performing search on the functional structure of the decoder. The decoder determines which pixel or voxel in a discretized solution is filled. Data-driven \textit{latent-generative} approaches use pre-trained trained generative models that compress a set of prior examples to a low-dimensional latent space, which serves as a search space $\mathbf{X}$.}
	\label{fig:encodings}
\end{figure}

We now discuss various encodings (see Figure \ref{fig:encodings}), their qualities w. r. t. the requirements from Section~\ref{sec:encodings:requirements}, and the implications to \gls{FDA}. 

\subsubsection{Direct/parameterized }

Most common encodings are direct and parameterized, defining a solution $\mathbf{s}$ based on a vector of parameters that is decoded into a shape. 
The parameter space $\mathbf{X}$, shown in green in Figure \ref{fig:encodings} (top), usually consists of the coordinates and/or weights of deformation nodes around a base shape. 
A naive approach can be to directly control the shape's nodes.
Other examples are airfoils \citep[see][]{hicks1974assessment}, splines \citep[see][]{vicini1999airfoil}, or \gls{FFD} of a base shape \citep[see][]{sederberg1986free}. 
\citet{sarakinos2005exploring} showed that \gls{FFD} can better compress $\mathbf{X}$ in less dimensions than spline representations.
\textit{Reachability} in $\mathbf{S}$ is limited by the hand-designed decoding, applying a parameter tuple to the user-selected base shape. 
Limited preliminary understanding of the domain can make it hard to define a decoder in the first place.
In contrast, a high degree of experience might lead to a more conservative stance, using well-understood decoders even when innovation is sought after.
\textit{Validity} of shapes can be influenced by putting constraints on the parameters. 
Constraints can be easy to implement but a large effort might be required to understand how constraints affect the reachable region $\mathbf{R}$ . 
In \gls{FDA} we want to avoid constraints initially. 
If $\mathbf{X}$ contains large regions of invalid solutions, constraints might be necessary to allow \gls{FDA} to be useful. 
However, due diligence is necessary to avoid missing solutions we do want.
\textit{Searchability} of direct encodings is given by the simple tuple structure of their parameters. 
The low dimensionality of $\mathbf{X}$ makes the search problem easier.
\textit{Predicting} a parameterized solution's quality is well-understood. 
The encodings can be used in efficient optimization schemes like Bayesian optimization, which will be explained in Section~\ref{sec:efficiency}. 
For complex solution structures, a higher-dimensional $\mathbf{X}$ is necessary which might make the necessary \textit{effort} to design a proper encoding high or even intractable. 
It is easy to \textit{understand} the solutions in $\mathbf{S}$ that will be reached, but hard to understand what regions are missed and whether they contain interesting solutions.
However, once artifacts that solve the problem are found, transferring them to the real world is easier. 
The engineer can control transferability through the decoder directly.
Translating \textit{prior examples} into the encoding is usually done through shape matching.
Shape matching is a research area that focuses on matching a prior shape using an encoding with an optimization algorithm \citep[see][]{tai2008target}.


\subsubsection{Indirect/Developmental/Generative}

Indirect (also called developmental or generative) encodings (Figure \ref{fig:encodings}, center) perform search in the decoder space.
Both the structure and weights of a neural graph, a common representation of a decoder called \gls{CPPN} \citep[see][]{stanley2007compositional}, can be changed.
A solution in $\mathbf{S}$ is created by the \gls{CPPN} by receiving the locations of pixels or voxels and returning whether that location is part of the solution or not. 
\citet{gaier2015evolutionary} showed that \glspl{CPPN} can be used to express \gls{NURBS}, making indirect encodings more appropriate to be used in engineering.
A single change in a search dimension may lead to multiple qualitative changes of the solution \citep[see][]{lapok2020evolving}.
An indirect approach can be used to reduce the number of search dimensions necessary to describe a wide variety of shapes. 
Indirect encodings can outperform direct encodings due to the reduction of the dimensionality of $\mathbf{X}$ \citep[see][]{hotz2004comparing}.
\citet{kicinger2006evolutionary} and \citet{clune2011performance} showed that indirect encodings work especially well when problem regularity increases, as regularity can be modeled by simple activation functions in the neural representation.
Other interpretations include the use of a Fourier series to encode the genes of a closed curve. \citet{yannou2008indirect} used this method to create car silhouettes with a fixed-dimensionality $\mathbf{X}$. 
The decoder can generate shapes in any resolution, helping to overcome the problem of maximizing the \textit{reachable} solution region $\mathbf{R}$ while using a low-dimensional $\mathbf{X}$.
By combining only a few elements of the decoder, a wide variety of shapes can be reached, as was shown in \citet{clune2013upload}.
\glspl{CPPN} are hard to \textit{constrain}, due to the complex expression function, but efforts have been made by including a post-processing CPPN results \citep[see][]{collins2019comparing}.
Regularities, like symmetry, can be injected by adding Gaussian building block functions into the \gls{CPPN}.
\glspl{CPPN} are usually \textit{optimized} using non-gradient-based evolutionary algorithms that are able to handle the fact that they can change structure and the unavailability of gradients \citep[see][]{stanley2002evolving,stanley2009hypercube}. 
Successful steps have been taken to connect graph structures like \glspl{CPPN} to \textit{efficient} predictive models \citep[see][]{hildebrandt2015using,stork2017surrogate,gaier2018data2,stork2019improving,hagg2019prediction}.
\glspl{CPPN}, which can change their internal structure and require a forward pass of pixel or voxel coordinates through the graph network, makes it complicated to \textit{understand} the expression and humans usually have to rely on posterior analysis of the resulting solutions.
Again, similar to direct encodings, we need to use search algorithms to match \textit{prior examples} using the encoding.
\citet{clune2013upload} showed that \gls{CPPN} representations can be matched to shapes.




\subsubsection{Latent-generative}
\label{sec:encodings:latent}

More complex search spaces can make \gls{FDA} an inefficient or even infeasible task.
Indirect encodings are a very flexible method to the problem of defining decoders \citep[see][]{stanley2002evolving,stanley2009hypercube}.
However, $\mathbf{S}$ might be simply too vast, complex and contain too many locally optimal solutions for non-data-driven techniques to be efficient and effective.
Although shape matching is a viable method to find prior examples with indirect encodings \citep[see][]{clune2013upload}, the encodings themselves are not trained in a data-driven manner.
The advent of \glspl{GM} skips the step of evolving an encoding and uses highly efficient machine learning techniques to represent large numbers of prior examples instead.
Figure \ref{fig:encodings} (bottom) shows an example of such a \gls{GM}, a \gls{VAE} that consists of two neural networks \citep[][]{kingma2014autoencoding}. 
The first, the encoder, compresses high-dimensional shape sets to a low-dimensional (latent) representation. 
The second network, the decoder, decompresses the latent space back into the original high-dimensional shape. 
Training is performed using a loss function on the decoder's output shape, comparing it with the training input.
Regularization terms can be added to increase the regularity of the latent space.
Alternative interpretations of \gls{GM} are \gls{GAN}, which use similar networks but train them in an adversarial manner. 
Here, the decoder, called the generator, tries to come up with realistic examples that are not contained in the training set.
The \textit{encoder} is changed into a classifier called the discriminator, which is responsible for distinguishing real from generated examples.

The compact design space provides an excellent search space in \gls{FDA}. 
The decoder has the potential to capture more relevant design characteristics than those human designers would suggest~\citep[see][]{rios2021exploiting}.
The authors showed that \gls{GM} can even allow local geometry modification in a 3D car design problem.
\Glspl{GM} can interpolate between training shapes or find new combinations of those shapes to produce innovative, novel solutions.
In another example, \glspl{GAN} have been used to synthesize aerofoils as well \citep[see][]{wang2021airfoil}.

\textit{Reachability} with \glspl{GM} is an incredibly hard problem to analyze, depending on the dimensionality of the latent space, which is directly connected to the accuracy of the reproduced training shapes.
While the state-of-the-art in image generation produces a very realistic diversity of images, it is not easy to determine the exact structure of $\mathbf{R}$, except by trial-and-error or search in the latent space.
The \gls{GM} is constrained to produce interpolations or new combinations of training shapes.
At the same time, these constraints create blind spots in areas the data set does not cover. However, \gls{GM} might reach human blind spots, finding more innovative solutions through data-driven discovery.
\citet{hagg2021expressivity} showed that in some cases, direct encodings might outperform \gls{GM} in an \gls{FDA} setting, where the dimensionality of the parameterized and latent encodings was the same.
Especially the diversity of output in a multi-solution context was shown to be higher for direct encodings. 

Constraints are not easy to implement, as \gls{GM} will just learn a representation based on the data it is presented. 
Validity has to be injected by using a data set consisting of valid shapes.
\citet{bentley2022coil} show that constraints on the training data can enable learning latent representations that produce mostly valid shapes.
They use a \gls{GA} to create a valid data set, which shows that an initial data set is necessary, which can be intractable to create for expensive optimization problems.
Without a data set, a \gls{GM} cannot be trained, but in order to efficiently create the data set, expensive computational effort is needed to optimize the expensive problem, in order to find valid data points.
A posterior inclusion of constraints usually has to be accomplished through post-processing or by retraining or adjusting the model by example.
An example of such a retraining was shown by \citet{hagg2020deep}, where the user was able to deselect solutions offered by the \gls{GM}.

The latent space provides a dimensionality reduction that makes search more feasible. 
The space itself, if trained in a regularized manner, can be continuous and smooth, by adding priors on the latent distribution.
Examples have appeared in the very recent past that show we can \textit{predict} performance of points in the latent space of a \gls{GM} with efficient models, either neural networks \citep[see][]{gomez2018automatic} or \glspl{GP} \citep[see][]{tripp2020sample}. 
The technique is quite new and its effectiveness probably depends on the smoothness of the latent space.
The ability to \textit{understand} the decoder in latent encodings is part of the bigger question of understanding neural network models. 
Explainability has become its own subfield and does not limit itself to ``analysis-by-example''.
Disentangling the latent dimensions of a \glspl{VAE} makes it easier to understand the latent space for humans~\citep[see][]{mathieu2019disentangling}.
Some work has been done on \glspl{GM} that produce simulation-ready meshes \citet{rios2021point2ffd}.
This decreases the effort of integrating a \gls{GM} into real world design and production processes.

Inserting \textit{prior knowledge} about solutions is the entire idea of \gls{GM}.
By learning representations from data, algorithms are fed with known good solutions, including prior knowledge of good and acceptable solutions into the model, automatically.
The usually large data sets of known prior solutions have to be transformed into a digital form that serves as training input to \glspl{GM}. 
Examples of such transformations are scans of analogue objects or technical drawings that are translated into a discretized bitmap or voxel representation. 
These representations can be presented to the \gls{GM} during training.

\subsection{Discussion}

An overview of this section is presented in Table \ref{tab:encoding:conclusion}.
\begin{table}[h]
	\centering
	\begin{tabular}{p{0.08\linewidth}|p{0.13\linewidth}|p{0.13\linewidth}|p{0.13\linewidth}|p{0.13\linewidth}|p{0.13\linewidth}|p{0.13\linewidth}}
		\textbf{Paradigm} & \textbf{Reachability} & \textbf{Validity} & \textbf{Searchability} & \textbf{Predictability} & \textbf{Understanding} & \textbf{Prior}\\
		Direct & Limited by design & Controllable & + & + & + & +/-\\ 
		Indirect & +& Hard& No gradients& +/-& Hard& +/-\\ 
		Latent & Can be outperformed by direct encodings, depends on the training data& Hard to constrain& Potentially lower dimensionality& Possible& Hard, needs disentangling, active research field& Per design\\ 
	\end{tabular}
	\caption{Overview of encoding types and requirements.}
	\label{tab:encoding:conclusion}
\end{table}

Direct encodings are well-understood but have the potential to limit reachability of solutions.
Indirect encodings, although having the potential to create more innovative shapes, are hard to predict and therefore can make them less efficient in the engineering context.
Latent encodings have a large potential and are efficient, but their performance is constrained by the number of data points available a priori. Simultaneously, being able to derive an encoding only based on data can help us circumvent the problem of having to manually design an encoding and extract it from the data directly.



\section{Search}
\label{sec:search}
We now discuss the question of how to use encodings to create diverse solution sets. 
In \gls{FDA}, the goal is not so much to come up with a global, valid optimum, but to understand the diversity of solutions and provide a more open-ended frame of thought.
This can lead to efficiency problems, because a large part of $\mathbf{X}$ might be less interesting to the user.
The search might spend too much time producing invalid solutions or even never reach valid regions.
On the other hand, invalid solutions, can be accepted as stepping stones to valid regions \citep[see][]{gaier2019quality,nordmoen2021map}.
To create these solutions, we can employ optimization algorithms. 
In fluid dynamics and efficient \gls{FDA}, although we might indeed want to understand as much about the domain as possible, we usually observe domains under the light of some definition of optimality. 
Optimality is defined as the objective to minimize a function $f$ on a solution, which determines the quality of a solution.

Common objectives in fluid dynamics are to minimize the drag force or the amount of turbulence in a flow caused by a shape. 
In these cases we are not interested in all shapes that we can produce in a domain but at least those that perform well according to the fitness function.
A performance function $f(\mathbf{x})$ that formalizes the optimization goal is defined based on the objective.
The (unconstrained) optimization problem is therefore defined as follows:
\begin{equation} \label{eq:optimization}
	\mathbf{x}_{min} = \argmin_\mathbf{x} (f(\mathbf{x}))
\end{equation}
where $\mathbf{x}_{min}$ is the global minimizer (or location of the minimum) of the function $f(\mathbf{x})$. 
$f(\mathbf{x})$ is the fitness value at $x$. 
A parameter tuple $\mathbf{x}$ is mapped to a solution $\mathbf{s}$ via a commonly user-defined encoding function or method $e$. 
The encoding $e$ usually entails computer code that determines the output format of a solution.

The goal of most classical optimization algorithms is to find a single, optimal solution. 
This clashes with the goal of \gls{FDA} - which is to understand as much about a domain as possible. 
To determine how different solutions actually solve a problem requires generating a large variety of designs. 
We therefore adapt Equation~\ref{eq:optimization} to output a set of locally optimal solutions $\mathbf{X}_{min,loc}$, which can contain the global optimum $\mathbf{x}_{min}$. 

\begin{figure}
	\centering
	\includegraphics[width=0.7\linewidth]{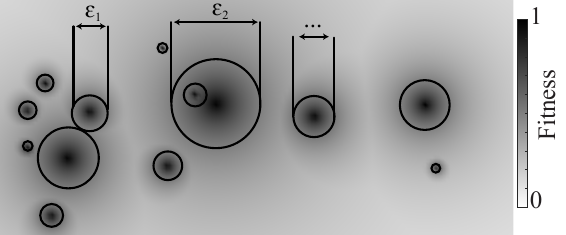}
	\caption{Heterogeneous fitness landscapes often contain clusters of varying sizes, making the definition of ``local optimum'' in terms of a threshold distance value $\epsilon$ indeterminable. $\epsilon_1$ and $\epsilon_2$ vary.}
	\label{fig:heterogeneousdomains}
\end{figure}

In order to produce $\mathbf{X}_{min,loc}$ we cannot merely define the fitness function and search for a minimizing (or maximizing) parameter tuple. 
We need to introduce a way to determine when we add a solution to the solution set. 
If we cannot make such a determination, the problem is malformed as we could produce any number of randomly selected solutions. 
Commonly, for solution selection, a criterion is used that takes the form of a threshold function, which would allow any solution below the threshold to be added to the solution set. 
The multi-solution optimization problem is defined as follows:
\begin{equation} \label{eq:multioptimization}
	\mathbf{X}_{min,loc} = \argmin_\mathbf{x} (f(\mathbf{x})), |\mathbf{x}_i - \mathbf{x}| \le \epsilon
\end{equation}
where $\mathbf{X}_{min,loc}$ is the set of solutions that minimizes $f(x)$ in a local neighborhood, a \textit{niche}.
A niche is defined by a distance-based metric $\epsilon$ on the parameters. 
$\epsilon$ is a domain-dependent parameter and might not be constant within the same domain, which can be observed in the heterogeneous fitness landscape in Figure \ref{fig:heterogeneousdomains}.
In equation \ref{eq:multioptimization}, $\epsilon$ also serves as a stand-in, as it might be determined explicitly, implicitly, or dynamically.

What follows are requirements of optimization algorithms used in \gls{FDA}.

\subsection{Requirements}

In the following sections, we define requirements to optimization algorithms and examples of methods, describe their strengths and weaknesses, and make recommendations for further research. 

\subsubsection{Multiple solutions}
\textit{The search should return multiple solutions.} To analyze an entire domain, optimization algorithms should only be considered if they return multiple solutions. Single-solution algorithms do not provide an overview of the solution space $\mathbf{S}$. The search algorithm should fulfill equation \ref{eq:multioptimization}.

\subsubsection{Coverage}
\textit{Search should maximize solution diversity.}
As we are interested in finding ``all'' solutions in a domain, the search method used in \gls{FDA} should cover as much of the solution space $\mathbf{S}$ as possible. 
As the search algorithm can only reach those points in $\mathbf{S}$ that are reachable by the encoding, we can simplify the requirement for search to maximize the coverage in $\mathbf{X}$. 

\subsubsection{Diversity}
\textit{Diversity of the solution set should be high.} 
In \gls{FDA} we are not only interested in finding as many solutions as possible or maximizing the spread in the search space.
Much more appropriately we want to create a diverse set of solutions that maximizes the knowledge gain of the user, and is a good representation for the variety of solutions.
Two aspects are of importance here: how we compare solutions and how well the space containing those solutions is sampled.
\citet{hagg2020analysis} discussed two aspects of diversity: the uniformity of the distribution of solutions (\textit{discrepancy}), spread (see last section) or a combination of the discrepancy and spread (\textit{coverage}).
Comparing solutions in $\mathbf{S}$ rather than $\mathbf{X}$ prevents genetic neutrality.
Neutrality is the phenomenon where multiple solutions in $\mathbf{X}$ are mapped to the same solution in $\mathbf{S}$.
Neutrality degrades diversity and diversity metrics \citep[see][]{hagg2020analysis}.
Within the space covered by search, solution diversity might be equally important to the user. 
For a review of diversity metrics, see \citep[][]{wang2016diversity}.

\subsection{State of the Art}

In multi-solution optimization we distinguish three candidate paradigms for \gls{FDA}, \gls{MOO}, \gls{MMO}, and \gls{QD}, each having a different idea behind the selection criterion.

\begin{figure}
	\centering
	\includegraphics[width=1\linewidth]{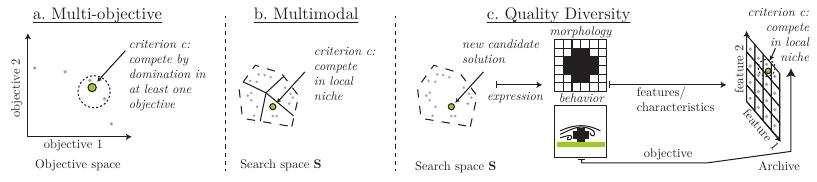}
	\caption{\textbf{Multi-solution optimization}. Multi-objective optimization (a) finds a Pareto front of trade-off solutions.
		Solutions are added to the front if they dominate neighboring solutions in at least one objective. 
		In multimodal optimization (b), solutions are selected through local competition in the parameter space. 
		Quality diversity (c) searches in parameter space, but local competition takes place in a low-dimensional archive defined by characteristics of their morphology or behavior.}
	\label{img:multisolution}
\end{figure}

\subsubsection{Multi-objective Optimization}

\Gls{MOO} (see figure \ref{img:multisolution}a) aims to produce different trade-offs between multiple optimization criteria. 
The selection criterion selects a solution when it performs better w. r. t. one of the optimization criteria.
The resulting Pareto front of solutions represents the trade-offs between multiple objectives.
Solutions are diverse w. r. t. this trade-off, yet not to their behavior or morphology.
The most successful methods to date are the \gls{NSGA-II}~\citep[see][]{Deb2002}, strength-Pareto evolutionary algorithm (SPEA)~\citep[see][]{zitzler2001spea2} and s-metric selection evolutionary multiobjective optimization algorithm (SMS-EMOA)~\citep[see][]{beume2007sms}.

\subsubsection{Multimodal Optimization}

\textit{Niching}, a concept from evolutionary optimization that protects solutions if they outperform close-by alternatives, is a concept that goes back to the 70s \citep[see][]{holland1975,Jong1975}. 
The idea was used to increase performance of single-objective evolutionary optimization first. 
Multisolution optimization came along almost two decades later.
The use of niching was now used to increase the number of output solutions of evolutionary algorithms.
Various algorithms have been introduced, like basin hopping~\citep[see][]{Wales}, nearest-better clustering~\citep[see][]{preuss2012improved} and restarted local search (RLS)~\citep[see][]{Posik2012}.

In \gls{MMO} (see figure \ref{img:multisolution}b), sets of solutions are created that spread out over the search (parameter) space and find as many (local) optima as possible.
Here, the selection criterion selects a solution when its location in $\mathbf{X}$ is far enough away from other known locations or its quality is higher of a close solution.

\subsubsection{Quality Diversity}

\Gls{QD} optimization solves the problem
\begin{equation} \label{eq:qd}
	\mathbf{X}_{min} = \argmin_\mathbf{x} (f(\mathbf{x})), |p(\mathbf{x}_i) - p(\mathbf{x})| \le \epsilon
\end{equation}
where $\mathbf{X}_{min}$ is the set of solutions that minimizes $f(\mathbf{x})$ in a local neighborhood, a \textit{niche}, defined by $\epsilon$ on the phenotypic characteristics considered by the function $p(\mathbf{x})$.

Finally, \gls{QD} (see figure \ref{img:multisolution}c) is a novel paradigm that aims to find a large diversity of high-performing solutions \citep[see][]{Hagg2020c}.
In \gls{QD}, diversity is defined in terms of the behavior or morphology of solutions, ignoring the diversity in parameter space altogether. 
In \gls{QD}, the selection criterion is similar to that of \gls{MMO}, except that locality is determined based on behavior or morphology, in $\mathbf{S}$, not in $\mathbf{X}$.
\gls{QD} was originally created to generate many walking gait strategies for a hexapod robot, allowing for quick restrategizing when the robot was damaged.
The behavioral characteristics describing the niching space were manually defined and well-suited for the purpose of generating strategies that used the robot's six legs in varying degrees.
Instead of manually defining features, which can be an intricate task,
\gls{QD} has been combined with latent-generative models~\citep[see][]{cully2019autonomous,gaier2020discovering,hagg2020deep,hagg2021expressivity}.
The generative models can either be used as a latent-generative search space (see section~\ref{sec:encodings:latent}), or as a feature/characteristic space, driving diversity of solutions.
The resulting feature models encode characteristics that can be used to define \gls{QD}'s archive dimensions. 
The drawback of such a purely data-driven approach is the lack of control over what solutions are found and how they are laid out to be compared to each other. 
A combination of data-driven and manually defined characteristics has not been researched in any work to the knowledge of the authors of this work.

\subsection{Comparison of Paradigms}

The three paradigms were analyzed and compared by \citet{hagg2020analysis}. 
The analysis showed that the diversity of solution sets produced by the three paradigms differs greatly.
\gls{QD} results in the highest diversity of solutions, making it an appropriate optimization method for \gls{FDA}. 
Although \gls{MOO} has been widely used to create multiple solutions in one go, solution diversity is only defined based on the trade-off between objectives.
This trade-off does not address diversity in terms of how a solution solves a problem, making it less appropriate for \gls{FDA} in early design phases.
While \gls{QD}'s resulting diversity is highest among the paradigms, \gls{MMO} does find a less diverse but higher-performing set and thus can be a viable alternative.
\gls{QD} is more effective on protecting solutions that are novel, but possibly less well-performing.
The trade-off between performance and diversity of a solution set is something to bear in mind when constructing an \gls{FDA} system.

Table \ref{tab:search:conclusion} summarizes this section. 
\begin{table}[h]
	\centering
	\begin{tabular}{c|c|c|c}
		\textbf{Paradigm} & \textbf{Coverage} & \textbf{Diversity} & \textbf{Applicability}\\
		\gls{MOO} & No, Pareto front & objectives, low & competing features\\ 
		\gls{MMO} & Yes (par.) & parameters, higher fitness & no features\\
		\gls{QD}  & Yes & features, higher diversity & behavioral features\\
	\end{tabular}
	\caption{Overview of multi-solution paradigms and their requirements.}
	\label{tab:search:conclusion}
\end{table}

When competing features, objective, exist, \gls{MOO} is a fitting paradigm. However, since in \gls{FDA} we are more interested in solution diversity, either \gls{MMO} or \gls{QD} are better paradigms. \gls{QD} produces the highest diversity, with the trade-off of including lower-performing individuals.

\section{Computational Fluid Dynamics}
\label{sec:simulation}

Computational Fluid Dynamics is a challenging field on its own; coupling it with FDA further increases complexity. Diverse shape sets produce qualitatively different flow conditions and requirements. On the one hand, the simulations must be accurate enough to capture the influence of tiny geometry changes on the optimized metrics. On the other hand, hundreds or thousands of unsupervised simulations need to be performed, i.e., the algorithm must be highly stable.  

\subsection{Level of detail}

The majority of realistic and engineering flows is subject to high Reynolds numbers and therefore turbulence. There are many different approaches when simulating turbulent flows, with implications to the full domain analysis framework, which is discussed in this paper. Due to the large number of simulations required, direct numerical simulations (DNS), resolving all relevant length and time scales, are too expensive for more realistic applications \citep[see][]{Alfonsi2009}, especially in three dimensions. The same mostly holds for the use of large-eddy simulation (LES), where low-pass filtering of the Navier-Stokes equations results in equations for the filtered large-scale quantities with unresolved subgrid terms, requiring modeling \citep[see][]{Sagaut2006}. This greatly reduces the simulation time and makes LES oftentimes feasible even for industrial applications \citep[see][]{Lohner2019}. Nevertheless, complex three-dimensional LES simulations are still too expensive to be used with FDA in all but limited selected cases. Statistical modeling, i.e. using the Reynolds averaged Navier-Stokes equations (RANS) \citep[see][]{Alfonsi2009} is still the most feasible approach for most optimization problems, especially when considering three-dimensional realistic applications.

\subsection{Stability vs. accuracy}
For standard CFD applications only a limited amount of simulations is required for a given problem, contrary to FDA, where the number of simulations is orders of magnitude larger. Special care needs to be taken regarding stability vs. accuracy. Practically, robust numerical algorithms and methods are preferable over less robust ones, even if those would provide higher accuracy. In FDA a vast number of unsupervised
simulations with different shapes need to be performed, therefore robustness is paramount. Otherwise, it is possible that interesting shapes are discarded because
of instabilities. 

Another example where reliability and accuracy must be weighed are meshing operations. The advantage of unstructured meshes is that the mesh can accurately capture the surface of the simulated object. On the flip side, meshing operations can easily consume more computational resources than the actual CFD simulation  \citep[see][]{Ingram2003}. In addition, low quality meshes can degrade the solution \citep[see][]{Mavriplis1997}. This can be an issue since the mesh generation in FDA is usually unsupervised. To counteract, mesh metrics can help identify and correct problematic mesh locations \citep[see][]{Balan2022}. A viable, fast alternative for FDA are also Cartesian grids with possible local grid refinements combined with cut-cell methods to account for complex boundaries \citep[see][]{Ingram2003}.

\section{On the Matter of Efficiency}
\label{sec:efficiency}
With appropriate encodings, we hope to reach a large or at least representative subset of $\mathbf{S}$. 
Multi-solution search methods introduced in section \ref{sec:search} are capable of producing it.
However, promising approaches like \gls{MMO} and \gls{QD}, require many, at least 100s or even 1000s, of simulations.
\gls{CFD} methods introduced in section \ref{sec:simulation} often need many hours in expensive 3D domains.
Efficiency therefore of utmost importance in order for \gls{FDA} to succeed.
In this section, we discuss the state of the art in efficiency enhancements for \gls{FDA} methods. 
Two strategies are distinguished: \textit{reducing} the number of necessary simulations by using clever sampling strategies and predicting flow behavior with cheaper models and \textit{replacing} simulations with a cheaper model altogether.

\subsection{Reduction}

This section has an overlap with \gls{SAO} \citep[see][]{jin2011surrogate}, where surrogate models serve as a cheap alternative to replace some of the real evaluations of individual solutions in simulation. 
In most cases, \gls{SAO} is used for single- or multiobjective cases. 
However, due to the large expected diversity in \gls{FDA}, surrogate models now have to predict characteristics and quality of a much more diverse solution set.
\citet{Wessing} gives some evidence that some \gls{SAO} ``instead has its strength in a setting where multiple optima are to be identified''. 

Using surrogate models, \gls{BO} can efficiently be used to discover diverse sets of optimal regions in expensive fitness domains \citep[see][]{gaier2018data,hagg2020designing}.
The two methods introduced there, SAIL and SPHEN, use the support of statistical models to efficiently predict what sampling locations are most effective at increasing information gain for the \gls{QD} optimization method.
Evidence was given that surrogate models are able to learn to predict flow characteristics simultaneously with predicting fitness.
The number of necessary real simulations was reduced by three order of magnitude to 1000, which started to make \gls{QD} methods feasible in expensive fluid dynamics domains. 
Surrogate-assistance can be applied in conjunction with indirect encodings \citep[see][]{gaier2018data2,hagg2019prediction}.

\subsection{Replacement}
In shape optimization the characteristics that are used to determine similarity and diversity of solution sets might be less easy to determine than in the original robotics cases of \gls{QD} literature \citep[][]{cully2015robots}.
Instead of relying on the prior knowledge (and biases) of engineers, data-driven techniques can be used for discovery of appropriate characteristics.
Deep generative models such as variational autoencoders (VAE) by \citet{kingma2014autoencoding} can extract patterns from raw data, learn meaningful representations for the data set. 
In combining \gls{QD} with latent-generative models~\citep[see][]{cully2019autonomous,gaier2020discovering,hagg2020deep,hagg2021expressivity}, representations can be developed that only produce high-performing solutions.
The disadvantage of these representations often is that the latent spaces are hard to interpret.
To alleviate this problem, disentangled representation learning can equip a model’s latent space with separated factors of variation revealing the underlying characteristics \citep[see][]{burgess2017understanding}.
The combination of disentangled representations and \gls{QD} seems to be a natural pairing, especially when one wants to understand correlations between aspects of morphology and flow without prior assumptions.
In order to use \gls{GM} in \gls{BO} settings, \citet{antonova2020bayesian} showed that the latent space of the \gls{GM} models can be used as a basis for surrogate assistance.

A wholly different approach is not to touch the encoding and instead predict the flow field directly using a deep neural network~\citep[see][]{chen2021numerical,wu2020deep,lye2020deep}.
Attempts are even made to train deep neural networks without any sampling data, by constraining networks with prior knowledge from physics ~\citet{sun2020surrogate}.

Other surrogate-assisted techniques use predictive models to connect coarse to fine models, e.g. multifidelity optimization \citep[][]{forrester2007multi} and space mapping \citep[][]{koziel2008space}.

\section{Full domain analysis demonstration}
\label{sec:example}

The encodings and algorithms (especially \gls{QD}) presented in the last sections together are able to generate large solution sets in an efficient manner. This section discusses how we present these results to the user, how \gls{FDA} can help the user to learn from data, how they can influence \gls{QD} by selecting shapes and how this leads to a hierarchical decomposition of a domain.
The user has to be able to explore a domain without being overwhelmed by the large amount of solution data.
The following requirements are necessary to allow in-depth analysis of such data sets:
\begin{itemize}
	\item Concise representation of diverse results
	\item Compact comparison
	\item Constrain by selection
	\item Change perspective
\end{itemize}

An example of an expensive domain is given where we efficiently create a large set of high-performing solutions, analyze their features, have a user zoom into a region of interesting solutions and study morphological features' correlation to flow features.


\subsection{Example Domain}

To show the possibilities \gls{FDA} gives the user, a simple 2D flow problem around spline shapes is constructed.
The domain, flow around 2D building footprints, is close to real world problems in fluid dynamics for the built environment, in which building norms put restrictions on the wind nuisance around buildings \citep[see][]{NEN8100,hagg2020designing}.
Wind nuisance is determined based on the maximum flow velocity around a building in typical wind conditions.
Flow around 2D shapes requires relatively little computational performance to simulate and is well understood.
Taking the role of the designing engineer during the computer aided design phase, we answer questions like what shapes lead to high levels of turbulence, is it possible to relate turbulence intensities and maximum flow velocity, and what morphological features cause high maximum velocity.

\begin{figure}[htb]
	\centering
	\includegraphics[width=1\linewidth]{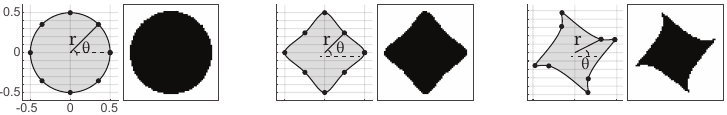}
	\caption{Encoding of 2D shapes: eight control points' polar coordinates.}
	\label{img:encoding}
\end{figure}

A 2D flow problem is constructed, with shapes inserted into that flow. 
The shapes are to induce a low (maximum) flow velocity $\mathbf{u}_{max}$ in the flow field.
The shapes are encoded as natural cubic splines, defined by eight control points, and then transformed into a 64x64 bitmap for evaluation (see Fig.~\ref{img:encoding}. 
The bitmaps are used in the \gls{LBM} solver \textit{Lettuce}~\citep[][]{lettuce} to calculate 2D flow around it. 
A Mach number of 0.075 and Reynolds number of 3900 was used in simulation.
2D footprints of high-rise building designs are evolved w. r. t. the same fitness function. 
Minimizing $\mathbf{u}_{max}$, high-rise buildings can be compliant with building regulations that prohibit strong gusts of wind around buildings in the built environment. 
We are interested in two features of solutions.
The area $\mathbf{A}$ of the footprint serves as a user-defined morphological feature of the domain along which solutions are varied. 
The second feature, the enstrophy $\mathcal{E}$, serves as a metric for the turbulence in the flow.
Of course we expect higher $\mathbf{u}_{max}$ to lead to higher $\mathcal{E}$, but this allows us to investigate whether we can learn this correlation from the (optimization) data.

\subsection{Search}

A diverse set of footprints is generated using \gls{SPHEN}. 
Diverse niches of solutions are created around the features area $\mathbf{A}$ of the footprint and enstrophy $\mathcal{E}$, turbulence. 
\gls{SPHEN} generates a Voronoi archive of 1000 solutions efficiently, using only 1000 \textit{Lettuce} samples. 
The number of samples can be reduced, of course, as can the archive size, for more expensive simulations. 
By using parallel evaluation processes, we can evaluate multiple \textit{Lettuce} instances at a time.

An initial sampling set of 100 2D shapes is generated by a pseudo-random sampling using a Sobol sequence \citep[see][]{sobol1967distribution} to generate 16-dimensional parameter tuples that describe those shapes.
We evaluate the shapes in \textit{Lettuce} to obtain $\mathbf{u}_{max}$ and $\mathcal{E}$.
The features have to be modeled in order to efficiently generate large solution sets with \gls{QD}.
Internal \gls{GP} surrogate models are trained to predict these flow features based on the shapes' parameters. 
The GP models use the isotropic squared exponential covariance function, as described in Eq.~\ref{eq:covariance}, to estimate the influence of samples on locations requested for prediction,

\begin{equation}
	k(x,x') = \sigma^2 \cdot exp\left(-\frac{(x - x')^2}{2l^2}\right)
	\label{eq:covariance}
\end{equation}

The covariance function's hyperparameters, length scale $l$ and signal variance $\sigma$, are determined by minimizing the log-likelihood using exact inference method of the GPML library \cite{rasmussen2010gaussian}.

\gls{SPHEN} uses the \gls{GP} models as surrogates for the expensive features to efficiently discover a diverse set of shapes with low $\mathbf{u}_{max}$.
The solutions are saved into a two-dimensional archive.
The archive is defined by the area $\mathcal{A}$ of the shape and $\mathcal{E}$. 
Newly generated solutions are assigned to the archive, if the archive is not full or if it outperforms the nearest neighbor in the archive.
The archive's maximum size is set to 1,000 solutions.
On every archive update 25 new solutions are created. 
New solutions are created by perturbing randomly selected solutions with a tuple pulled from a normal distribution with $\sigma = 0.1$.
After updating the archive 1,000 times, having created and compared 25,000 new solutions in total, we end up with an archive of 1,000 (predicted) high-performing solutions.
Ten new samples are then selected, again using a Sobol sequence to ensure they are spread out in the archive.
These samples are now evaluated in \textit{Lettuce}, to obtain new data for the \gls{GP} models, which are retrained after every round of the internal \gls{QD} search.
This process is continued until we have obtained 1,000 shape samples with the accompanying features, as determined by \textit{Lettuce}.
\begin{figure}[htb]
	\centering
	\includegraphics[width=1\linewidth]{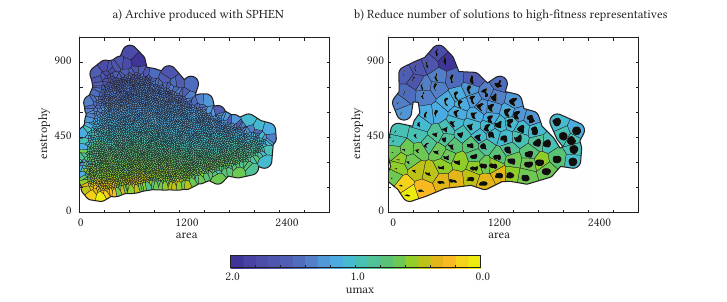}
	\caption{After efficiently training surrogate models for the features (area and enstrophy/turbulence) and fitness of a diverse solution set, we can easily produce an archive of 4000 solutions (a) and reduce the solution set size to its best 100 representatives (b).}
	\label{img:archives}
\end{figure}
Although we used 1,000 expensive simulations, we were able to evaluate, in a surrogate-assisted manner, 2,250,000 proposed solutions.
After producing 1000 samples, the \gls{GP} models can predict $\mathbf{u}_{max}$, $\mathbf{A}$, and $\mathcal{E}$ for a diverse set of solutions. 
Fig.~\ref{img:archives} shows the archive produced with \gls{SPHEN}. 

\subsection{QD Analysis Step}
The user can now analyze the shapes, their features and correlations to fitness. 
Users can select high-fitness representative shape examples (Fig.~\ref{img:archives}b). \gls{SPHEN} does not have to be fully reevaluated to achieve this, except rebuilding the archive with a smaller size.
As we can already see from this archive, the lowest $\mathbf{u}_{max}$ is reached when the area $\mathbf{A}$ and enstrophy $\mathcal{E}$ are small.
However, a trade-off appears, as the larger $\mathbf{A}$ becomes, the higher $\mathbf{u}_{max}$.
A linear correlation between $\mathcal{E}$ and $\mathbf{u}_{max}$ is visible as well, as we expected.

\subsection{Generation Step}

Although we used a fixed encoding to generate the shape archive, we now have \gls{GP} models that allow us to create larger archives.
A new archive with 4000 solutions is trained and a \gls{VAE} is trained based on this data set.
A larger data set can be created with ease, but not necessarily in this use case.
The \gls{VAE} will allow us to generate new solutions, interpolate between solutions and find disentangled morphological features that help us understand the correlations between shape and flow features. 

\begin{figure}[htb]
	\centering
	\includegraphics[width=1\linewidth]{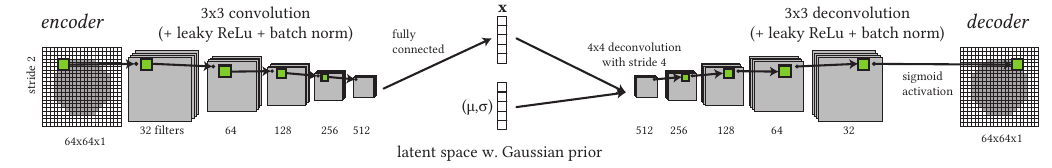}
	\caption{Architecture of the convolutional \gls{VAE} generative model.}
	\label{img:cvae}
\end{figure}

Figure \ref{img:cvae} shows the used architecture of the convolutional \gls{VAE}. 
The filters, with size 3x3 and stride 2, reduce the resolution of the input image by a factor of 2.
Every convolutional layer contains more filters, to hierarchical decompose (bitmaps of) the shape set into ever smaller, more low-level, morphological features. 
This encodes the shape into a five-dimensional space of latent variables that describe core morphological aspects of the shape set.
Latent variables are comparable to principal components, except that they are non-linear.
To decode the latent variables back into a shape, deconvolutions are used in the exact yet mirrored order as in the encoder.

\subsection{VAE Analysis Step}

The \gls{VAE}'s smooth latent space offers a new morphological search space that produces only shapes with relative low $\mathbf{u}_{max}$.
Flow values can be predicted using \gls{GP} models on the basis of latent coordinates of shapes.
The \gls{VAE} allows the user to vary morphological features around a selected shape. 

\begin{figure}[htb]
	\centering
	\includegraphics[width=1\linewidth]{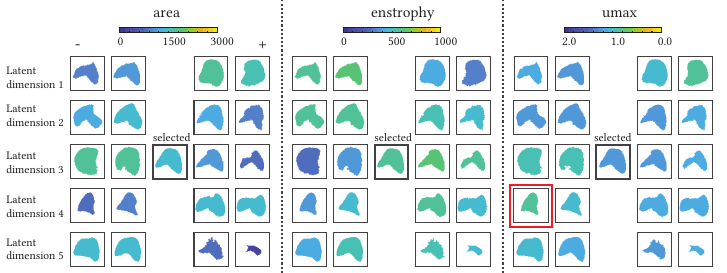}
	\caption{The three features are shown for a shape and its variants, when varying five latent dimensions in a \gls{VAE} model. Red marks an interesting shape with a lower $\mathbf{u}_{max}$ than the originally selected one (center column).}
	\label{img:latentwalk}
\end{figure}

Figure \ref{img:latentwalk} shows what happens when each of the five latent variables is varied (rows).
The color indicates the value of the predicted $\mathbf{u}_{max}$, $\mathbf{A}$ and enstrophy $\mathcal{E}$.
The user can analyze how morphological changes influence predefined morphological features like $\mathbf{A}$ or flow features.
As can be seen from the figure, $\mathcal{E}$ and
$\mathbf{u}_{max}$ are positively related. 
The higher $\mathcal{E}$, the higher $\mathbf{u}_{max}$, as expected.

\begin{figure}[htb]
	\centering
	\includegraphics[width=0.7\linewidth]{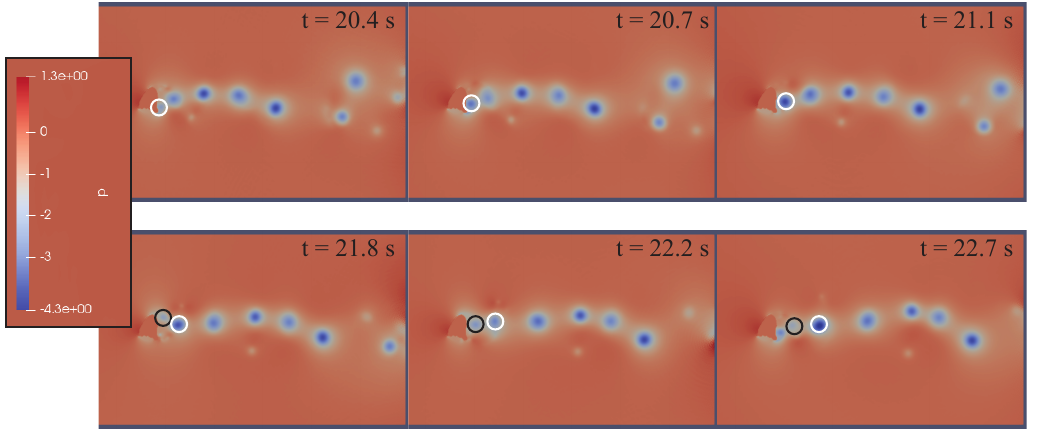}
	\caption{Flow around selected shape. The top row shows the appearance of a strong vortex (white outline) caused by the bottom of the shape. The bottom row shows a weak vortex (black outline) caused by the top of the shape, appearing after the strong vortex.}
	\label{img:flow}
\end{figure}

The user might now select a shape generated by the \gls{VAE} for further studying. The shape selected, marked in red in figure \ref{img:latentwalk}, has a lower $\mathbf{u}_{max}$ than the original shape. Now, the user chooses to run an actual simulation to validate the predicted flow features and understand, why the shape has a lower $\mathbf{u}_{max}$. Figure \ref{img:flow} shows six consecutive time stamps of the flow produced with \textit{Lettuce}. Vortices appear at the bottom and top side of the view successively, shown in white and black outlines. The actual value of $\mathbf{u}_{max} = 2.49$ is higher than predicted. 

\subsection{Very Large Solution Sets}

\begin{figure}[htb]
	\centering
	\includegraphics[width=1\linewidth]{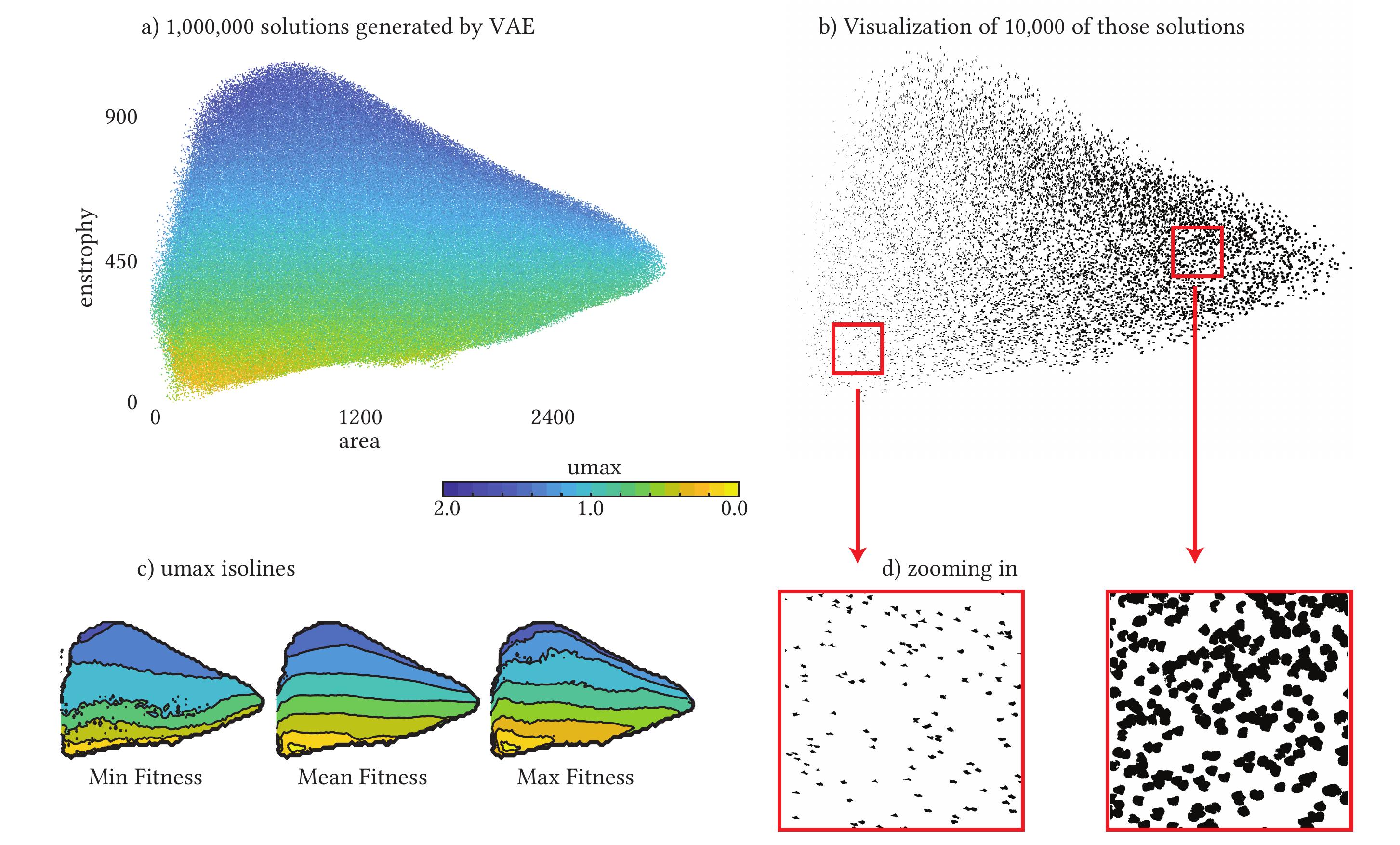}
	\caption{1,000,000 solutions generated by \gls{VAE} (a), subselection of 10,000 shapes (b), isolines of min, mean and max fitness (c), and zoomed in shape regions (d). }
	\label{img:largearchive}
\end{figure}

To really take \gls{FDA} to its extremes, we generated 1,000,000 solutions, shown in Figure \ref{img:largearchive}.
It can be clearly seen that $\mathcal{E}$ and $\mathbf{u}_{max}$ have an almost linear relationship (Figure \ref{img:largearchive}a).
However, the relationship is not purely linear, as can be seen from the minimum, mean, and maximum fitness isolines (Figure \ref{img:largearchive}c).

The user can zoom in on specific regions of interest to perform a more in-depth analysis of local morphological effects (Figure \ref{img:largearchive}d). 

\begin{figure}
	\centerline{\includegraphics[width=1\textwidth]{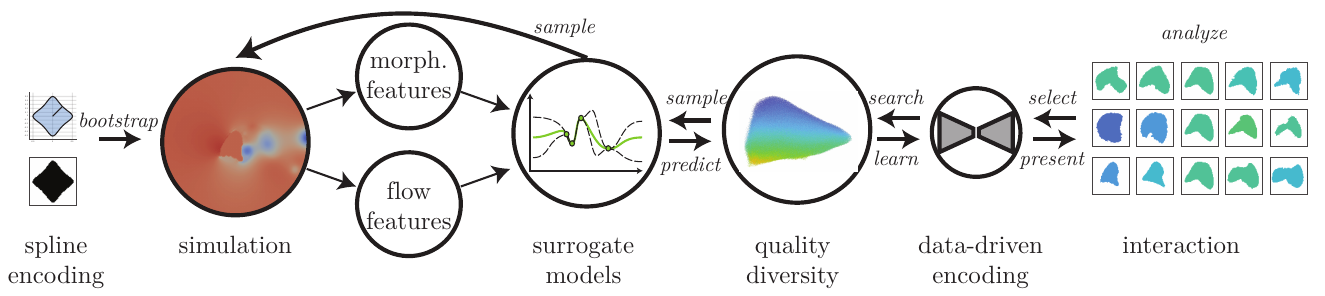}}
	\caption{\gls{FDA} implementation example. First, the \gls{QD} process is bootstrapped with a predefined spline encoding. Using surrogate models to assist with predicting and sampling examples based on their quality and diversity, \gls{QD} produces an archive of the solution space that is presented to the user. A data-driven encoding can be trained based on the shapes from the \gls{QD} archive. Using this encoding and the archive, the user can analyze the solution space. They can then zoom in on a region of that space, analyze data-driven morphological features, their correlation to flow features and interact with the system. The data-driven encoding both compresses the search space, can serve for further investigation, search and provide new diversity metrics to be fed back into the \gls{QD} process.}
	\label{fig:frameworkExample}
\end{figure}

Finally, the entire example \gls{FDA} system is shown in Figure \ref{fig:frameworkExample}. By using a classical encoding to bootstrap a quality diversity algorithm, which efficiently samples a fast \gls{CFD} simulation, it is possible to produce a large, diverse set of instances or shapes. 
A data-driven encoding is then developed to generate even more shapes, allowing the user to analyze morphological dimensions in the latent space as well as perform other in-depth analyzes.
The following forms of interaction between the user and the system can be distinguished: analysis of morphological space, morphological and flow features, generating variations of shapes, selecting shapes and constraining the \gls{QD} process in further iterations.

\section{Conclusions}
\label{sec:conclusions}


The ability to efficiently understand and analyze the full space of solutions and their behavior provides an innovative support tool for engineers.
Possible answers to the questions asked in the introduction (Section \ref{sec:introduction}) were given in this article, but we also outlined avenues for future research.
We showed that direct encodings can provide us with a bootstrapping mechanism for \gls{FDA} in fluid mechanics.
\gls{FDA} efficiency is increased using statistical models that are able to predict both the quality of solutions as well as their similarity. The models are used to guide the sampling method towards a diverse set of optima.
Surrogate-assisted \gls{QD} provides an efficient sampling method to find high-performing solutions based on either a manually defined encoding, for bootstrapping, or a data-driven encoding. 
A GPU-based \gls{LBM} solver was used to determine quality and diversity/similarity of solutions. 
Although other CFD solvers may also be suited, this particular \gls{LBM} solver is tuned towards automated flow simulations around diverse shape sets.
The results can be returned to the user, who can analyze the results efficiently and influence the sampling. 
A latent model can be trained on the resulting set, providing a bootstrapped generator of high quality solutions and a means to visualize and understand how morphological features are correlated to flow features.
The user then needs to be able to ``zoom in'' on certain solution classes on-demand. 
The interactivity aspect of the framework is highlighted in other work \citep[][]{hagg2020deep}. 

Improvements of the presented ingredients of \gls{FDA} are many, since the respective fields are developing quickly.
Novel \gls{QD} methods might improve the efficiency of the search even more. Examples of these methods are the integration of a faster derivative-free optimizer \cite{fontaine2020covariance}, adding differentiability when it is available \cite{fontaine2021differentiable}, exchanging the sampling function \cite{kent2020bop}, or further integration of deep learning methods \cite{zhang2021deep}, are just some of the examples where the search itself can be made more efficient and effective.
The \gls{GM} can be improved further by keeping track of research on disentangling the latent dimensions, making them more interpretable for humans.

New insights often lead to new focuses in the analysis of complex domains. 
The current development in evolutionary algorithms and machine learning enables us to more deeply understand problem domains, even when they are as expensive as fluid dynamics.

\section*{Acknowledgments}
The authors would like to thank Lea Prochnau and George Khujadze for their contributions. 
\section*{Funding}
This work received funding from the German Federal Ministry of Education and Research (BMBF) under the ``Forschung an Fachhochschulen mit Unternehmen'' programme (grant agreement number 03FH012PX5).
The computer hardware was supported by the Federal Ministry for Education and Research and by the Ministry for Innovation, Science, Research, and Technology of the state of Northrhine-Westfalia (research grant 13FH156IN6).

\bibliographystyle{apalike}
\bibliography{references,references_AHAG,references_fluid-dyn}


\end{document}